\documentclass[11pt]{article}
\usepackage{coling2020}
\usepackage{times}
\usepackage{url}
\usepackage{latexsym}
\usepackage[dvipdfmx]{graphicx}
\usepackage{xspace}
\usepackage{amsmath,amsthm,amssymb,ascmac}
\usepackage{algorithm}
\usepackage{algorithmicx}
\usepackage{algpseudocode}
\usepackage{tikz}
\usepackage{xcolor}

\def\checkmark{\tikz\fill[scale=0.4](0,.35) -- (.25,0) -- (1,.7) -- (.25,.15) -- cycle;}

\makeatletter
\newcommand{\figcaption}[1]{\def\@captype{figure}\caption{#1}}
\newcommand{\tblcaption}[1]{\def\@captype{table}\caption{#1}}
\makeatother

\newcommand\bertbase{BERT$_{\small \textsc{BASE}}$\xspace}

\title{Diverse and Non-redundant Answer Set Extraction on Community QA based on DPPs}
\author{Shogo Fujita$^1$\thanks{~~This work was done as an intern at Yahoo Japan Corporation.}, Tomohide Shibata$^2$ and Manabu Okumura$^1$ \\
  $^1$Tokyo Institute of Technology \\
  $^2$Yahoo Japan Corporation \\
  {\tt \{fujisyo,oku\}@lr.pi.titech.ac.jp} \\
   \tt tomshiba@yahoo-corp.jp
  }
\date{}

\begin{document}
\maketitle
\begin{abstract}
\blfootnote{
    \hspace{-0.65cm} 
    This work is licensed under a Creative Commons 
    Attribution 4.0 International Licence.
    Licence details:
    \url{http://creativecommons.org/licenses/by/4.0/}.
}

In community-based question answering (CQA) platforms, it takes time for a user to get useful information from among many answers. Although one solution is an answer ranking method, the user still needs to read through the top-ranked answers carefully.
This paper proposes a new task of selecting a diverse and non-redundant answer set rather than ranking the answers.
Our method is based on determinantal point processes (DPPs), and it calculates the answer importance and similarity between answers by using BERT.
We built a dataset focusing on a Japanese CQA site, and the experiments on this dataset demonstrated that the proposed method outperformed several baseline methods.

\end{abstract}

\section{Introduction}
Community-based question answering (CQA) platforms have been used by many people to solve problems that cannot be solved simply by searching in books or web pages.
On a CQA platform, the questioner posts a question and waits for others to post answers.
The posted answers are viewed by not only the questioner herself but also other users.

A single question may receive several to dozens, even hundreds, of answers, and it would take a lot of time to see all the answers.
The questioner chooses the most helpful answer, called the best answer, and users can view this answer as a representative answer.
However, there may be more than one appropriate answer to a question that asks for reasons, advice, and opinions, while there is often only one appropriate answer for a question that asks for facts, etc.  
For example, Table~\ref{sample} shows example answers to the question {\it ``Why do birds turn their heads backwards when they sleep?''}, which asks for reasons.
The best answer A1 says it is to keep out the cold and A4 says it is for balance, which is also an appropriate answer.
In such a case, if users view only the best answer, they miss other appropriate answers.

One possible way to deal with this problem is to use ranking methods~\cite{10.1145/1571941.1571974,10.1145/1390334.1390416,10.1145/1367497.1367561,10.5555/3298023.3298081,10.1145/3077136.3080790,10.5555/3298023.3298080,laskar-etal-2020-contextualized}.
Since the studies use the best answers as training data, similar answers are ranked higher although the existing methods try to reduce redundancy by using the similarity between a question and an answer or between answers.
As a result, users need to read through carefully the answers that are ranked higher to obtain the information they want to know.

We propose a new task of selecting a diverse and non-redundant answer set from all the answers, instead of ranking the answers.
We treat the set of answers as the input, from which the system determines both the appropriate number of answers to be included in the output set and the answers themselves.
The datasets used in the conventional ranking task of CQA are not suitable for our task because they contain only a ranked list of answers, whereas we need to determine the number of answers to be included in the output set in addition to the answers themselves.
Therefore, we have constructed a new dataset for our task that contains diverse and non-redundant annotated answer sets.

We use determinantal point processes (DPPs)~\cite{macchi_1975,10.5555/2481023} to select a diverse and non-redundant answer set.
DPPs are probabilistic models that are defined over the possible subsets of a given dataset and give a higher probability mass to more diverse and non-redundant subsets.
To estimate the answer importance and the similarity between answers, we use BERT~\cite{devlin-etal-2019-bert}, which has achieved high accuracy in various tasks.

Focusing on a Japanese CQA site, we built training and evaluation datasets through crowdsourcing and  conducted experiments demonstrating that the proposed method outperformed several baseline methods.
\begin{table}[t]
 \small
  \resizebox{\columnwidth}{!}{
  \begin{tabular}{cp{43em}|c}
\hline
   Q&\multicolumn{2}{l}{Why do birds turn their heads backwards when they sleep?} \\
    \hline
    \hline
    &\multicolumn{1}{c|}{answer} & gold\\
    \hline
    \textbf{A1} & \textbf{...Because it's cold. They keep their beaks in their feathers \color{red}{to protect them from getting cold.}}&\checkmark \\
    \hline
    A2&This is because it reduces the surface area of the body \textcolor{red}{to prepare for a drop in body temperature and to prevent physical exhaustion.}&\\
    \hline
    A3&\textcolor{teal}{Well balanced.}& \\
    \hline
    A4&They look back \textcolor{teal}{to stabilize their bodies} by shifting their center of gravity on the vertical line of their feet to stabilize their bodies.&\checkmark \\
    \hline
    A5&I think it's to use their feather as pillows.&\\
\hline
  \end{tabular}
  }
  \vspace{-3mm}
  \caption{Example of a question and its answers where multiple answers are considered to be appropriate. The answer in bold is the best answer. Sentences in the same color indicate they have the same content. Gold indicates answers included in the reference answer set.}
  \label{sample}
\end{table}

\section{Related Work}
\subsection{Work on CQA}

CQA has been actively studied~\cite{nakov-etal-2017-semeval,nakov-etal-2016-semeval-2016,nakov-etal-2015-semeval}. The most relevant task to our study is \textit{answer selection}.

There are two approaches to the answer selection task.
One approach is to predict the best answers~\cite{tian-towards-2013,7314643}.
The other is to rank the entire list of potential answers~\cite{10.1145/1571941.1571974,10.1145/1390334.1390416,10.1145/1367497.1367561,10.5555/3298023.3298081,10.1145/3077136.3080790,10.5555/3298023.3298080,laskar-etal-2020-contextualized}.
The problem with these ranking methods is that they do not take account of the similarity between answers.
To solve this problem, \newcite{10.1145/2911451.2911506} proposed a method for ranking answers based on the similarity between questions and answers and between answers, and \newcite{10.1145/3308558.3313457} proposed a method for ranking answers using a neural model that was learned to select answers that are similar to the best answer and less similar to the answers already selected.
These studies are related to our method in that they take account of diversity in selecting from the answers in CQA.
Moreover, they address the task of selecting one appropriate answer or sorting them in appropriate order, which is different from our task.

A number of methods of summarizing answers in CQA ~\cite{10.1145/3297001.3297004,10.1145/3018661.3018704} also take diversity into account.
These methods extract sentences of a fixed length, and therefore, cannot determine an appropriate output length.
Our method is different in that it can determine the appropriate output length by using DPPs.
Some methods of question retrieval in CQA also take diversity into account~\cite{10.1145/1772690.1772712,Szpektor2013}.

\subsection{Work on Diversity}

Taking account of diversity is important in information retrieval and summarization, and methods using maximal marginal relevance (MMR)~\cite{10.1145/290941.291025} have been proposed~\cite{boudin-etal-2008-scalable,10.1145/1835449.1835639}.

DPPs~\cite{macchi_1975,10.5555/2481023} has also been proposed as a way for taking account of diversity.
This method performs better than MMR in these fields.
DPPs have been applied to recommendation problems as a way of taking diversity into account~\cite{10.1145/3269206.3272018,10.5555/3327345.3327465,Mariet2019DPPNetAD}.
Multi-document summarization models combine DPPs and neural networks to take account of diversity ~\cite{cho-etal-2019-improving,cho-etal-2019-multi}.
DPPs have been also used in studies on clustering verbs~\cite{reichart-korhonen-2013-improved} and extracting example sentences with specific words \cite{tolmachev-kurohashi-2017-automatic}.

\section{Proposed Method}
\begin{figure}[t]
\begin{center}
\includegraphics[width=40em]{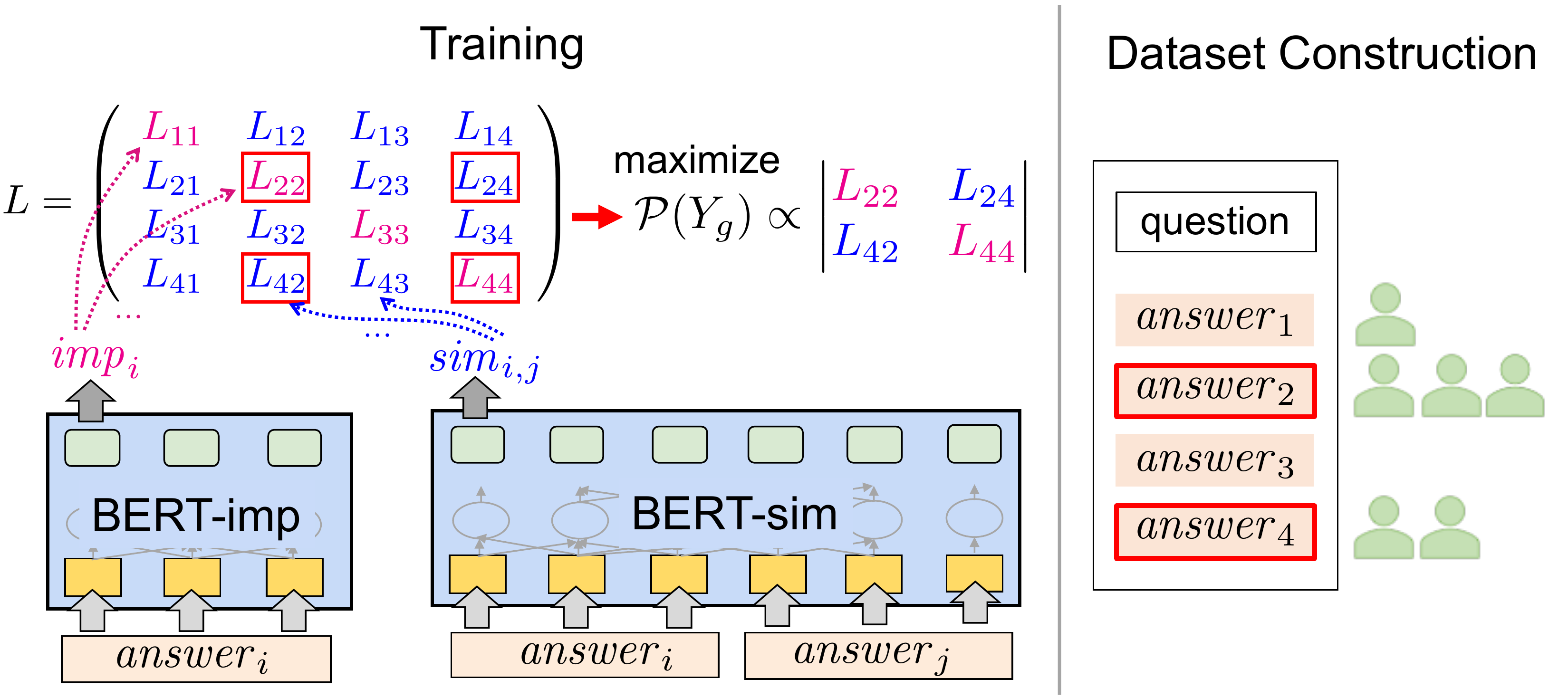}
 \vspace{-3mm}
 \caption{Overview of the proposed method.}
 \vspace{-4mm}
 \label{model}
 \end{center}
\end{figure}
The overview of our method is shown in Figure~\ref{model}.  
Our method uses DPPs to select a diverse and non-redundant subset from a set of answers. The DPPs assign a higher probability to a subset in which each answer is important and the answers are diverse. BERT~\cite{devlin-etal-2019-bert} is used for estimating the importance of an answer and the similarity between answers. In the example shown in Figure~\ref{model}, answers 2 and 4 are annotated as a gold answer set. The probability of choosing these gold answers is defined by the DPPs. Intuitively, BERT is fine-tuned so that the importance of answers 2 and 4 is higher and the similarity between answers 2 and 4 is lower.

\subsection{DPPs}
DPPs define the probability measure $\mathcal{P}$ over all subsets ($2^N$) of a ground set containing all the answers $\mathcal{Y}=\{1,2,...,N\}$ where $N$ denotes the number of answers.
Our method computes the probability of choosing a subset of answers based on the kernel matrix $L\in \mathbb{R}^{N\times N}$ that contains the importance of the answers and the similarities between answers, and it selects the subset $Y\subseteq \mathcal{Y}$ with the highest probability.
The probability measure $\mathcal{P}$ is defined as
\begin{align}
    \mathcal{P}(Y;L)&=\frac{det(L_Y)}{\sum_{Y'}{det(L_{Y'})}}\\
    &=\frac{det(L_Y)}{det(L+I)},
\end{align}
where $L_Y\in \mathbb{R}^{|Y|\times |Y|}$ is a submatrix containing only
entries indexed by elements of $Y$, $I\in \mathbb{R}^{N\times N}$ is the identity matrix, and $det(\cdot)$ is the determinant of a matrix. 
The kernel matrix $L$ is a positive semi-definite matrix.
The $i,j$-th component of the kernel matrix $L_{i,j}$ is defined as
\begin{align}
L_{i,j}=imp_i \times imp_j \times sim_{i,j},
\end{align}
where $imp_i$ is the importance of the $i$-th answer and $sim_{i,j}$ is the similarity between the $i$-th and $j$-th answers.
The similarity of the same answer $sim_{i,i}$ is set to 1.
From the definition of the determinant, the DPPs select important answers and do not select similar answers at the same time.
Following \newcite{10.1145/3269206.3272018}, the kernel matrix is decomposed into eigenvalues, and negative eigenvalues are replaced with tiny values to satisfy the semi-positive definiteness constraint.\footnote{While \newcite{10.1145/3269206.3272018} replace the negative eigenvalues with zero, we replace them with tiny values to stabilize the learning.}
In the training, the parameters of BERT are updated so that the following negative log-likelihood is minimized:
\begin{align}
-log&(\mathcal{P}(Y_g;L))=-log(det(L_{Y_g}))+log(det(L+I)),
\end{align}
where $Y_g \subseteq \mathcal{Y}$ is the gold answer set. 

At inference time, our method computes the probability of all the candidate answer sets and selects the one with the highest probability.
Note that since the dataset used in our experiments had a relatively small number of answers, we did not need to adopt an approximate method by sampling.

\subsection{BERT Architecture}
The architecture has two mechanisms based on BERT: BERT-imp, which uses the BERT mechanism to calculate the answer importance, and BERT-sim, which calculates the similarity between answers. 
Following the BERT paper, the \texttt{[CLS]} and \texttt{[SEP]} tokens are added to the beginning and end of a sequence of tokens, respectively.

\subsubsection{BERT-imp}

BERT-imp is a mechanism for predicting the importance of an answer.
$imp_i$ is calculated as follows: 
\begin{align}
&imp_i=\exp({\rm MLP}^{imp}({\rm BERT}_{CLS}^{imp}({\rm input}^{imp}(a_i)))),\\
&{\rm input}^{imp}(a_i)=\texttt{[CLS]}a_i\texttt{[SEP]},    
\end{align}
where $a_i$ is the $i$-th answer, and ${\rm BERT}_{\rm CLS}^{imp}$ is a function that returns the hidden state vector corresponding to \texttt{[CLS]} by vectorizing the input using the BERT mechanism.
${\rm MLP}^{imp}$ is a feed-forward network with a single hidden layer that outputs the importance of an answer.
Since DPPs are log-linear models, 
the $\exp$ function is applied after the output of the last layer.

\subsubsection{BERT-sim}
BERT-sim is a mechanism to compute the similarity between answers.
Following standard practice, we separate the two answers with \texttt{[SEP]} tokens.
$sim_{i,j}$ is calculated as follows: 
\begin{align}
&sim_{i,j}=\sigma({\rm MLP}^{sim}({\rm BERT}_{\rm CLS}^{sim}({\rm input}^{sim}(a_i,a_j)))),\\
&{\rm input}^{sim}(a_i,a_j)=\texttt{[CLS]}a_i\texttt{[SEP]}a_j\texttt{[SEP]},    
\end{align}
where ${\rm BERT}_{\rm CLS}^{sim}$ is the same mechanism as ${\rm BERT}_{\rm CLS}^{imp}$, and
${\rm MLP}^{imp}$ is the same as ${\rm MLP}^{sim}$.
The $\sigma$ function is applied to make the value in the range $[0,1]$.

Our method is similar to the DPPs-based summarization model~\cite{cho-etal-2019-multi}. \newcite{cho-etal-2019-multi} use a single document summary dataset to train BERT-imp, and BERT-sim and the feature weight for BERT-imp is learned during DPP training.
In our method, because we did not have the appropriate data to pre-train BERT-imp and BERT-sim, 
BERT-imp and BERT-sim are fine-tuned in an end-to-end fashion using only the training data for the target task.

\section{Experiments}

\subsection{Dataset Construction}
\label{subsec:dataset_construction}

Since there was no dataset for the task of selecting a diverse and non-redundant answer set in the CQA field, we constructed one for this study.
We focused on Yahoo! Chiebukuro,\footnote{https://chiebukuro.yahoo.co.jp/} the largest CQA service in Japan.
From among categories with many questions, we excluded categories, in which almost all the answers are different, such as cooking, and categories, in which most of answers are too long, such as human relationships, and we selected questions in the healthcare, diet, and pet categories. 
We excluded questions and answers that were too short and uninformative in each category; in particular, we excluded those questions whose best answers were less than 110 characters in length and those questions with less than five answers.

We used a crowdsourcing service called Yahoo! Crowdsourcing for annotation.
For a total of approximately 10,000 questions in the above three categories, the questions and answers were presented to 10 workers, and they were asked to select a diverse and non-redundant answer set.
When crowdsourcing is used, only instances that most workers agree on should be chosen for training and evaluation datasets.
In our task, the chosen answer set may vary from person to person, and the number of the chosen answers tends to be small (usually one).
As such, simply choosing only the questions that many workers agree on would be undesirable because the dataset would be small, and the number of answers in almost every gold answer set would be one.

\begin{figure}[t]
    \def\@captype{table}  
  \begin{minipage}[c]{.44\textwidth}
    \begin{center}
    \small
    \begin{tabular}{c|c|c|c|c}
    &\multicolumn{4}{l}{ \begin{tabular}{c}\# of workers who chose\\ the same answer set\end{tabular}}\\
\hline
        & 4&3&2&1 \\\hline
        A1 & \textcolor{red}{\checkmark}&\textcolor{red}{\checkmark}&\checkmark&\checkmark\\\hline
        A2 & &\textcolor{red}{\checkmark}&&\\\hline
        A3 & &&\checkmark&\\\hline
        A4 & &&&\checkmark\\\hline
        A5 &~~~~~~~~&~~~~~~~~&~~~~~~~~&~~~~~~~~\\\hline
    \end{tabular}
    \end{center}
        \begin{center}
    \vspace{-3mm}
    \tblcaption{Example of annotation results.}\label{select-exp}
    \resizebox{\columnwidth}{!}{
        \begin{tabular}{c|c|c|c|c|c|c}  \hline
        &\multicolumn{6}{c}{\# of answers in gold answer sets}\\\hline
        & 1&2&3&4&5&6 \\\hline
        \# of questions&2552&923&155&34&10&6\\\hline
        ratio &0.69&0.25&\multicolumn{4}{c}{0.06}\\
\hline
    \end{tabular}
    }
    \vspace{-3mm}
    \tblcaption{Distribution of the number of answers in gold answer sets.}\label{gold}
    \end{center}
  \end{minipage}
    \begin{minipage}{.56\textwidth}
    \begin{center}
  \includegraphics[width=\linewidth]{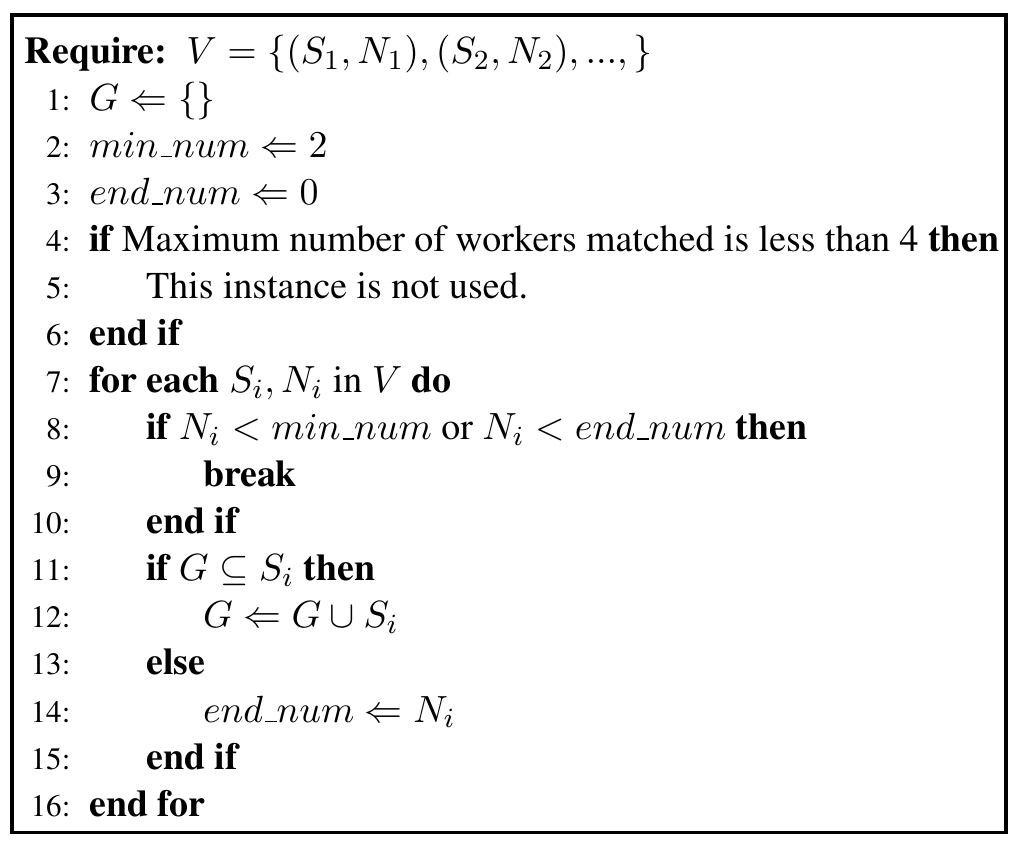}
  \vspace{-9mm}
  \caption{Algorithm for selecting the gold answer set $G$.}\label{alg1}
  \end{center}
  \end{minipage}
\end{figure}

Therefore, we loosened the concept of ``matching'' and devised an algorithm that allows multiple answers to be selected as much as possible.
Specifically, the algorithm first chooses only those instances for which there is some degree of agreement in the annotation results.
The algorithm searches in order from the answer set chosen by most workers.
If the answer set includes a current gold answer set that has already been chosen, the algorithm adds it to the current gold answer set.
We refer to Table~\ref{select-exp}, which shows a typical annotation result, to illustrate the behavior of the algorithm.
The most popular choice is \{A1\}, so the algorithm first adds it to the current gold answer set.
\{A1, A2\} chosen by three workers includes the already chosen current gold answer set \{A1\}, so the algorithm adds A2 to the current gold answer set.
Since \{A1, A3\} chosen by two workers does not include the already chosen current gold answer set \{A1, A2\}, the algorithm stops. 
The algorithm outputs \{A1, A2\} as the gold answer set.

The detailed algorithm is shown in Figure~\ref{alg1}.
$V$ is an aggregated list of votes from workers for each selected answer set.
For each aggregate, $S$ represents the answer set and $N$ represents the number of workers who chose $S$.
$V$ is sorted in descending order of $N$, and if $N$ is the same for two or more items on the list, it is sorted in ascending order of the number of answers in $S$.
The $i$-th element of $V$ is $(S_i,N_i)$.
In the case of the example shown in Table~\ref{select-exp}, $S_1 = \{A1\}$, $N_1 = 4$, $S_2 = \{A1, A2\}$, and $N_2 = 3$.

If the answer set chosen by $k$ workers does not include the current gold answer set, the algorithm should look at all the answer sets chosen by the other $k$ workers before terminating.
For this reason, we use a variable $end\_num$ indicating the number of workers who chose the last answer set that the algorithm would see.
To consider only the answer sets selected by two or more workers, the constant $min\_num$ is set to be $2$.

Table~\ref{non-select-exp} shows an example of a non-selected question.
Both A3 and A5 contain information that it is important to get into the habit of training every day and not to overdo it.
Although there are minor differences, the answers are similar.
In this case, three workers chose A3 and three other workers chose A5, which caused a split in opinion.
Such questions were not included in the dataset because of the difficulty in determining an appropriate answer set.

The number of questions in the dataset is 3,680.
Table~\ref{gold} shows the distribution of the number of answers in the gold answer sets.
The average number of the answers is 1.38. 
While 69\% of the questions have one answer in the gold answer sets, 6\% of the questions have three or more answers.
Our dataset was randomly split into a training set (2,576 instances), development set (368 instances), and testing set (736 instances).

\begin{table}[t]
 \small
 \resizebox{\columnwidth}{!}{
    \begin{tabular}{cp{33em}|c|c|c|c|c}
\hline
       Q&\multicolumn{6}{l}{ \begin{tabular}{l}I've heard that abdominal exercises, lying on a back and lifting and lowering legs, \\ are effective in reducing the amount of fat in the lower abdomen. \\I've done too much and my muscles are sore, should I take a break from this?\end{tabular}}\\\hline \hline
    &\multicolumn{1}{c|}{answer} &  \multicolumn{5}{c}{ \begin{tabular}{c}\# of workers who chose\\ the same answer set\end{tabular}} \\\hline
      &&3&3&2&1&1\\\hline
    A1&In my case, when I have muscle pain, I'll \textcolor{teal}{keep doing it}...&&&&&\checkmark\\
    A2&\textcolor{red}{Don't overdo it}, but \textcolor{teal}{don't take a break.}&&&&&\\\hline
    A3&\textcolor{teal}{It's important to continue every day.} \textcolor{red}{The key is to continue in moderation and without strain}...\textcolor{orange}{If it hurts too much, you can take a break.}&\checkmark&&\checkmark&\checkmark&\\\hline
    A4&I'm not sure which muscle is sore, but if it's not your stomach, it doesn't have much to do with your lower abdomen, does it?...&&&&\checkmark&\\\hline
    \textbf{A5}&\textbf{\textcolor{orange}{How about working your muscles other than your abdominal muscles?} ...\textcolor{teal}{It's important to get into the habit of exercising every day} \textcolor{red}{rather than pushing yourself in the beginning}...}&&\checkmark &\checkmark&\checkmark&\\\hline
  \end{tabular}
 }
 \vspace{-4mm}
  \caption{Example of the instance that was not chosen in our dataset.}
  \vspace{-2mm}
  \label{non-select-exp}
\end{table}
\subsection{Experimental Settings}

Accuracy was used to assess the overall agreement, while precision, recall, and f1 were used to assess partial agreement.
Our method used \bertbase, which was pre-trained in the Whole Word Masking setting from Japanese Wikipedia.\footnote{https://github.com/cl-tohoku/bert-japanese}
The number of dimensions of the hidden layers of ${\rm MLP}^{imp}$ and ${\rm MLP}^{sim}$ was set to 100.
Following the BERT paper, AdamW was used as the optimizer.
The learning rates (2e-5, 3e-5, and 5e-5) used in the BERT paper were found to be rather large for our task, so thus a smaller one, 2e-6, was used instead.
We trained and evaluated each model three times for 5 epochs and computed the average scores of the models with the highest accuracy on the development set.

Since our task is one of selecting an appropriate answer set, our method cannot be directly compared to the previous studies where the appropriate number of answers in an answer set is unknown. Therefore, the following baseline methods determine the importance of an answer, and select the fixed number of answers for all the questions. Two variants for each baseline method were considered: one outputting the most important answer, the other outputting the two most important answers, since the average number of gold answer sets is 1.38.
Hereafter, the method that chooses one in a method M is called M-1, and the method that chooses two is called M-2.
The following baselines were compared with our proposed method:
\\
\itemsep=0pt
\textbf{random} As a simple baseline, we experimented with a model to select answers randomly.\\
\textbf{wordnum} 
Since a longer answer has a lot of information, it is likely included in a gold answer set.
This baseline selects an answer in descending order of its word number.\\
\textbf{lexrank} LexRank \cite{10.5555/1622487.1622501} is a graph-based method for text summarization, which calculates the importance of a sentence by taking account of the similarity between sentences. This method was applied to our task. This baseline selects an answer in descending order of LexRank score.\\
\textbf{ranking} 
To compare with the conventional ranking method~\cite{10.1145/3308558.3313457}, we ran a model to select answers ranked by that method.
This model requires that one appropriate answer be chosen in advance.
Therefore, we ran a model to select the best answer using BERT.
We used 
approximately 55k questions to learn the model to choose the best answer.
For training on the dataset, we constructed five input sequences, each containing the question and the answer\footnote{If a question had more than five answers, five answers were randomly sampled, including the best answer.}.

The model calculated the importance of each answer using the hidden state vectors corresponding to \texttt{[CLS]} and selected the answer with the highest score.
\noindent

To verify the effectiveness of BERT, we experimented with a model that used bidirectional long short term memory (bi-LSTM)~\cite{Hochreiter:1997:LSM:1246443.1246450} for the encoder (\textbf{LSTM}).
The LSTM encoder outputs the concatenation of the last hidden states of forward and backward LSTM.
We replaced the BERT-based encoders used by BERT-imp and BERT-sim with bi-LSTM.
The word embeddings were initialized using pre-trained word2vec embeddings~\cite{DBLP:journals/corr/abs-1301-3781,10.5555/2999792.2999959}. The word2vec embeddings were pre-trained using Japanese Wikipedia.
The optimizer was AdamW, and a learning rate of 5e-4 was chosen from among \{1e-3, 5e-4, 1e-4\} by using the development set.

\subsection{Results}
\begin{table}[t]
\centering
      \begin{tabular}{c|c|ccc}
\hline
Model&Acc.&Prec.&Rec.&F1\\
\hline\hline
     random-1& 0.141&0.262&0.193&0.222\\
     random-2& 0.022&0.257&0.378&0.306\\
     wordnum-1&0.395&0.643&0.473&0.545\\
     wordnum-2& 0.129&0.495&\textbf{0.728}&0.589\\
     lexrank-1& 0.337&0.565&0.416&0.479\\
     lexrank-2&0.098&0.450&0.662&0.536\\
     ranking-1&0.325 &0.561 &0.413 &0.476 \\
     ranking-2& 0.056& 0.385& 0.567& 0.459\\
\hline
     LSTM & 0.418&0.672&0.495&0.570\\ 
\hline
     Proposed &\textbf{0.477}&\textbf{0.719}&0.582&\textbf{0.643}\\ 
\hline
  \end{tabular}
  \vspace{-3mm}
    \caption{Evaluation results for the test set.}\label{result}
    \vspace{-3mm}
\end{table}

Table~\ref{result} shows the results for the test set.
Among the baseline methods, the models that chose one answer achieved higher accuracy and precision, and the models that chose two answers achieved higher recall and F1.
In terms of accuracy, \textbf{wordnum-1}, which selects the answer with the most words, performed the best among the baseline methods.
The result is considered reasonable, as longer answers are more likely to contain more information.

\textbf{Proposed} outperformed the simple baselines and \textbf{ranking}, and its results demonstrated that the approach of creating a dataset for selecting a diverse and non-redundant answer set and learning on DPPs is effective.
\textbf{Proposed} also outperformed \textbf{LSTM}, which is consistent with recent studies showing that pre-trained models are effective.

\subsection{Analysis}

We performed several analyses to better understand how our method works.
\paragraph{Adding a question to BERT-imp} 
In BERT-imp introduced in Sec 3.2.1, the input is an answer.
Not only the answer but also the question likely contributes to the prediction of the answer's importance.
For this reason, a method that adds a question to the input in BERT-imp was tested~(\textbf{+Q\_BERT-imp}).

\begin{figure}[t]
  \def\@captype{table}
  \begin{minipage}{.50\linewidth}
      \begin{center}
    \resizebox{\columnwidth}{!}{
      \begin{tabular}{c|c|ccc}
\hline
     Model& Acc. &  Prec. & Rec.& F1\\
\hline\hline
      Proposed &\textbf{0.441}&0.723&\textbf{0.579}&\textbf{0.643}\\
\hline
      +Q\_BERT-imp&0.440&\textbf{0.734}&0.569&0.641\\
      +BA\_pretrain & 0.411&0.697&0.550&0.615\\
\hline
     BA &0.408&0.704&0.522& 0.600\\
\hline
  \end{tabular}
  }
  \end{center}
  \vspace{-3mm}
    \tblcaption{Evaluation results used in the analysis. These scores were measured with the development data.}\label{result2}
  
  \end{minipage}
\begin{minipage}[c]{.50\linewidth}
    \begin{center}
\includegraphics[width=0.9\linewidth]{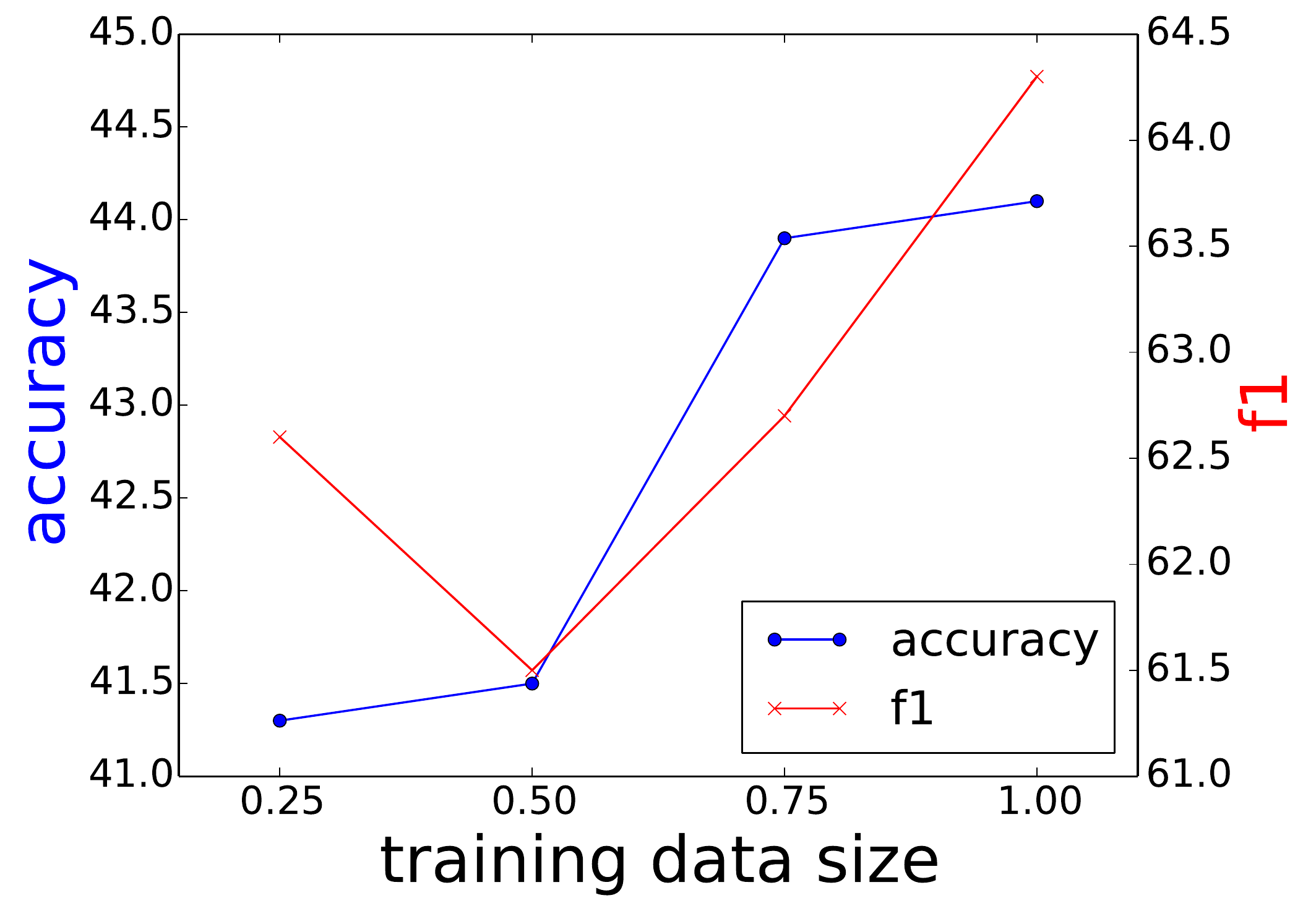}
\vspace{-5mm}
\caption{Correlation between the training data size and the metrics.}
 \label{train_num}
     \end{center}
  \end{minipage}
\end{figure}

The upper-half of Table~\ref{result2} shows the results.
\textbf{+Q\_BERT-imp} performed almost as well as \textbf{Proposed}, which demonstrates that the question did not contribute to the prediction of the answer importance.
Since our method cannot capture the relevance of questions and answers with a small corpus, it is considered sufficient for BERT-imp to focus only on answers.

Table~\ref{que-dif} shows an example of different outputs between \textbf{Proposed} and \textbf{+Q\_BERT-imp}.
While A3 and A4 are the same in that they both recommend manila grasses as the easiest lawn to maintain, they have different contents about planting methods, so it is appropriate to choose both.
The reason why \textbf{+Q\_BERT-imp} only chose A3 as the correct answer set to this question is that the model rated the importance of A3, which is similar to the question, higher, while it rated the importance of A4, which included a lot of content not directly related to the question, lower.
The dataset contains answers that are not appropriate answers to the question, such as A2 in this case; answers such as these can be judged by looking only at the answers.

\paragraph{Pre-training for BERT-imp} 
The task of choosing the best answer might be relevant to our task.
To compensate for the paucity of annotated data, we can pre-train BERT-imp in the task of choosing the best answer.
This pre-training setting is the same as the model for choosing one answer in advance used in \textbf{ranking} in Sec 4.2.
We ran the model which uses an additionally pre-trained BERT-imp ~(\textbf{+BA\_pretrain}).

The bottom-half of Table~\ref{result2} shows the results.
\textbf{+BA\_pretrain} had lower accuracy and F1 than \textbf{Proposed}, which demonstrates the pre-training in the task of choosing the best answer prevented a prediction of answer importance.

To test the correlation between the best answer and the gold answer set, we measured the scores in the setting of choosing only the best answer (\textbf{BA}).
The percentage of the best answer in the gold answer set is 0.704, so the best answer might not be included in the gold answer set.
Since the best answer is chosen by the questioner alone, it is prone to biases and does not correlate strongly with the gold answer set.
The pre-training makes the model choose the answers that are most likely to be considered the best answers, and this may have led to a decrease in performance.

Table~\ref{pre-dif} shows an example of different outputs between \textbf{Proposed} and \textbf{+BA\_pretrain}.
A3 is the only answer in the gold answer set to the question.
In the answers to a question about which doctor to see if the questioner has cystitis, A3 answers that it doesn't matter whether questioner sees a physician or a gynecologist.
Although the gold answer set to this question only contains A3, \textbf{+BA\_pretrain} chose A4, which contains only information that it is OK to see a gynecologist and does not include information that it is no problem to see an internist.
Thus, the best answer tends to be sympathetic or positive to the questioner, so it is not always the most informative one.

\begin{table}[t!]
 \small
 \resizebox{\columnwidth}{!}{
    \begin{tabular}{cp{31em}|c|c|c} 
\hline
   Q & \multicolumn{4}{l}{ \begin{tabular}{l}I would like to plant a lawn. What is the easiest lawn to care for and maintain?\\ What is the best way to plant lawns? \end{tabular}}\\
\hline\hline
      &\multicolumn{1}{c|}{answer}&gold&Proposed&+Q\_BERT-imp\\
\hline
    A1&\textcolor{red}{The easiest lawn to maintain and care for is manila grass} that is easily purchasable. \textcolor{teal}{The key to planting is to level the ground} so that you can use an electric lawnmower.&&&\\
\hline
    A2&This isn't the answer, but you should know that managing your lawn can be a daunting task.&&&\\
\hline
    A3&...\textcolor{red}{An easy one is manila grass}, which is often sold at home improvement stores...\textcolor{teal}{To plant it, level the substrate} and \textcolor{brown}{add fertilizer to the subsoil beforehand}...&\checkmark&\checkmark&\checkmark\\
\hline
    \textbf{A4}&\textbf{Two types of turf are usually available on the market...The most important part of planting is to \textcolor{magenta}{loosen the soil to some degree}....\textcolor{violet}{Fertilizer is not necessary if the grass grows there}...\textcolor{olive}{Lightly sprinkling soil over the surface of the turf} at planting will help it to root better. \textcolor{orange}{Water should be sprayed thoroughly}...\textcolor{red}{Manila grass is the easiest to care for.} Zoysia japonica is the best looking.}&\checkmark&\checkmark&\\
\hline
    A5&Take care of any lawn to prevent it from being invaded by white clover.&&&\\
\hline
  \end{tabular}
  }
  \vspace{-5mm}
  \caption{Example of different outputs between \textbf{Proposed} and \textbf{+Q\_BERT-imp}.}
  \vspace{-2mm}
  \label{que-dif}
\end{table}

\begin{table}[t!]
 \small
 \resizebox{\columnwidth}{!}{
    \begin{tabular}{cp{32em}|c|c|c}
\hline
   Q & \multicolumn{4}{l}{ \begin{tabular}{l}I seem to have cystitis. I need to go to the hospital; should I see a urologist? \\ Or should I see a gynecologist?\end{tabular}}\\
\hline\hline
      &\multicolumn{1}{c|}{answer}&gold&Proposed&+BA\_pretrain\\
\hline
    A1&\textcolor{red}{Go to an internal medicine clinic.} It's better to see a specialist afterwards.&&&\\
\hline
    A2&...It doesn't matter which one you go to. &&&\\
\hline
    A3&\textcolor{red}{You can see a physician as well.} \textcolor{teal}{If you have a family gynecologist, that's fine too.} A gynecologist treats all kinds of women's diseases...
&\checkmark&\checkmark&\\
\hline
    \textbf{A4}& \textbf{I recently had cystitis too. That was painful. The gynecologist's questionnaire had a ``cystitis'' section on it, and that's when I first realized that \textcolor{teal}{gynecologists generally treat cystitis}...}&&&\checkmark\\
\hline
    A5&Talk to the hospital...&&&\\
\hline
  \end{tabular}
  }
  \vspace{-5mm}
  \caption{Example of different outputs between \textbf{Proposed} and \textbf{+BA\_pretrain}.}
  \label{pre-dif}
\end{table}

\paragraph{Varying the size of training data}
We investigated how the performance changes when the training data size varies. Figure \ref{train_num} shows the performance of \textbf{Proposed} when the training data size was set to $0.25$, $0.5$, or $0.75$ in the development set. The performance tended to be improved as the corpus size increased, and did not saturate. Therefore, increasing the training data size is expected to improve performance.

\subsection{Case Study}
 
\begin{table}[t!]
 \small
 \resizebox{\columnwidth}{!}{
  \begin{tabular}{cp{39em}|c|c}
\hline

  Q&\multicolumn{3}{l}{
   \begin{tabular}{l}Many black bugs stuck to the underside of the leaves of a potted lvy.\\ If possible, I would like to get rid of them without using insecticides.\\ If there is a good method, please let me know. \end{tabular}}\\
\hline\hline
    &\multicolumn{1}{c|}{answer} & gold & output \\
\hline
    A1&If it's an aphid, \textcolor{red}{spray it with milk.} If it's a spider mite, \textcolor{teal}{pour water on the back of the leaves} and you'll drown it. & &  \\
\hline
    A2&I think it would be a good idea to \textcolor{brown}{light some mosquito coils...} & & \\
\hline
    A3&I would \textcolor{blue}{use wood vinegar}...& \checkmark&\checkmark\\
\hline
    A4&Why do you not want to use pesticides?
\textcolor{orange}{Pesticides pose no danger when used correctly!}...& & \\
\hline
    \textbf{A5}&\textbf{...1) \textcolor{red}{Spray with milk}. 2) \textcolor{violet}{Strain and spray crushed garlic and pepper powder boiled down with soju.} 3) \textcolor{magenta}{Soak cigarette butts in water to extract the nicotine content, and spray.} It's not a plant to eat, and  \textcolor{orange}{wouldn't it be better to spray a pesticide to get rid of them in one shot?}...}& \checkmark&\checkmark\\
\hline
  \end{tabular}
  }
  \vspace{-5mm}
  \caption{Example of correct output.}
  \label{exp1}
\end{table}

\begin{table}[t!]
 \small
 \resizebox{\columnwidth}{!}{
  \begin{tabular}{cp{40em}|c|c}
\hline
   Q & \multicolumn{3}{l}{ \begin{tabular}{l}It is often said that as we age, our skin stops repelling water, but is it true?\\ Also, if possible, please tell me the cause. \end{tabular}}\\
\hline
\hline
    &\multicolumn{1}{c|}{answer}&  gold & output  \\
\hline
    A1&In my case, my towel got wet a lot more often...& & \\
\hline
    A2&Because you \textcolor{brown}{lose its elasticity}, you can't get polka dots on your skin...& &\\
\hline
    A3&I think \textcolor{brown}{it's a loss of elasticity}, or in other words, a loss of facial muscles. & &\\
\hline
    \textbf{A4}&\textbf{\textcolor{red}{It's true.}...\textcolor{teal}{The oil that my body produces on the surface of my skin becomes less and less secreted as I get older}, so it stops repelling water...In the case of the face, it becomes less oily, \textcolor{brown}{loses its elasticity}...} &\checkmark&\checkmark\\
\hline
    A5&...I've heard that on skin that has \textcolor{brown}{lost tension etc.,} the water droplets are not spherical, but instead is spread out in gooey droplets.……& &\checkmark\\
\hline
  \end{tabular}
  }
  \vspace{-5mm}
  \caption{Example of incorrect output.}
  \label{exp2}
\end{table}

We analyze some of the outputs of our method.
Table~\ref{exp1} shows an example of a correct output.
The method chose A3 and A5.
Both A3 and A5 contain the information about how to remove black bugs from the leaves, and both are included in the gold answer set.
A5 does not contain the information in A3 about using a wood vinegar, and A3 does not contain the information in A5 about spraying milk.
Therefore, the chosen answer set is non-redundant.
The contents of A1 and A4 were included in A5, and our method was able to extract the various information from the answers.

Table~\ref{exp2} shows an example of an incorrect output.
A4 is the most informative answer and includes the content of A5.
However, our method has chosen both A4 and A5.
As shown in this example, our method sometimes chose redundant answers, which may indicate insufficient learning of the similarity.
In the future, its performance could be improved by creating a dataset with similar answers and pre-training BERT-sim to estimate the similarity between answers.

\section{Conclusion}
We proposed a new task to select a diverse and non-redundant answer set in CQA.
We created a dataset for this task, and proposed a model for it by using DPPs and BERT.
Our method performed better than several baseline methods.
In future work, we aim to pre-train BERT-sim to calculate the similarity between answers and apply our method to CQA platforms in other languages.

\bibliographystyle{coling}
\bibliography{coling2020}

\end{document}